\documentclass[conference]{IEEEtran}
\IEEEoverridecommandlockouts
\usepackage{cite}
\usepackage{algpseudocode,algorithm,algorithmicx}
\usepackage{amsmath,amssymb,amsfonts}
\usepackage{amsthm}
\usepackage{graphicx}
\usepackage{booktabs}
\usepackage{textcomp}
\usepackage{xspace}
\usepackage{subcaption}
\usepackage{caption}
\usepackage{color}
\usepackage{multirow}

\usepackage{xcolor}
\usepackage{soul}

\usepackage{tabularx,booktabs}
\newcolumntype{C}{>{\centering\arraybackslash}X} 

\newcommand{\etal}{\textit{et al}.\@\xspace}

\newcommand{\volume}{\ooalign{$V$\cr\raisebox{0.15em}{\kern0.04em--}\cr}}

\begin{document}

\title{Pay ``Attention'' to Adverse Weather: \\
Weather-aware Attention-based Object Detection}

\author{\IEEEauthorblockN{Saket S. Chaturvedi}
\IEEEauthorblockA{\textit{College of Computing} \\
\textit{Michigan Technological University}\\
Houghton, MI, USA \\
schatur2@mtu.edu}
\and
\IEEEauthorblockN{Lan Zhang}
\IEEEauthorblockA{\textit{Dept of Electrical and Computer Engineering} \\
\textit{Michigan Technological University}\\
Houghton, MI, USA \\
lanzhang@mtu.edu}
\and
\IEEEauthorblockN{Xiaoyong Yuan}
\IEEEauthorblockA{\textit{College of Computing} \\
\textit{Michigan Technological University}\\
Houghton, MI, USA \\
xyyuan@mtu.edu}
}

\maketitle

\begin{abstract}
Despite the recent advances of deep neural networks, object detection for adverse weather remains challenging due to the poor perception of some sensors in adverse weather. Instead of relying on one single sensor, multimodal fusion has been one promising approach to provide redundant detection information based on multiple sensors. However, most existing multimodal fusion approaches are ineffective in adjusting the focus of different sensors under varying detection environments in dynamic adverse weather conditions. Moreover, it is critical to simultaneously observe local and global information
under complex weather conditions, which has been neglected in most early or late-stage multimodal fusion works. In view of these, this paper proposes a Global-Local Attention (GLA) framework to adaptively fuse the multi-modality sensing streams, i.e., camera, gated camera, and lidar data, at two fusion stages. Specifically, GLA integrates an early-stage fusion via a local attention network and a late-stage fusion via a global attention network to deal with both local and global information, which automatically allocates higher weights to the modality with better detection features at the late-stage fusion to cope with the specific weather condition adaptively. Experimental results demonstrate the superior performance of the proposed GLA compared with state-of-the-art fusion approaches under various adverse weather conditions, such as light fog, dense fog, and snow. 

\end{abstract}
\vspace{0.5em}

\section{Introduction}
Object detection is a fundamental task in autonomous driving. Reliable and robust detection is crucial to various downstream autonomous driving tasks, such as object tracking~\cite{manjunath2018radar}, path planning~\cite{minh2014feasible}, and motion control~\cite{sharma2019lateral}. Although recent advances in deep neural networks have significantly improved object detection performance in normal weather, their performance in adverse weather such as light fog, dense fog, and snow has been challenging \cite{zang2019impact}.

Object detection in adverse weather is significantly hindered by the poor performance of some sensors.
For example, cameras alone are unreliable in low-light conditions as they do not provide depth information and are highly affected by adverse weather ~\cite{zang2019impact}. Similarly, the effective range of lidar is highly impaired in fog and snow due to backscatter~\cite{heinzler2020cnn}. To provide robust perception in adverse weather, multimodal sensor fusion is one of the most promising approaches since it leverages the benefits from multiple sensors to provide redundant detection information.

\begin{figure}[t]
    \centering
    \includegraphics[width=8cm, height=5cm]{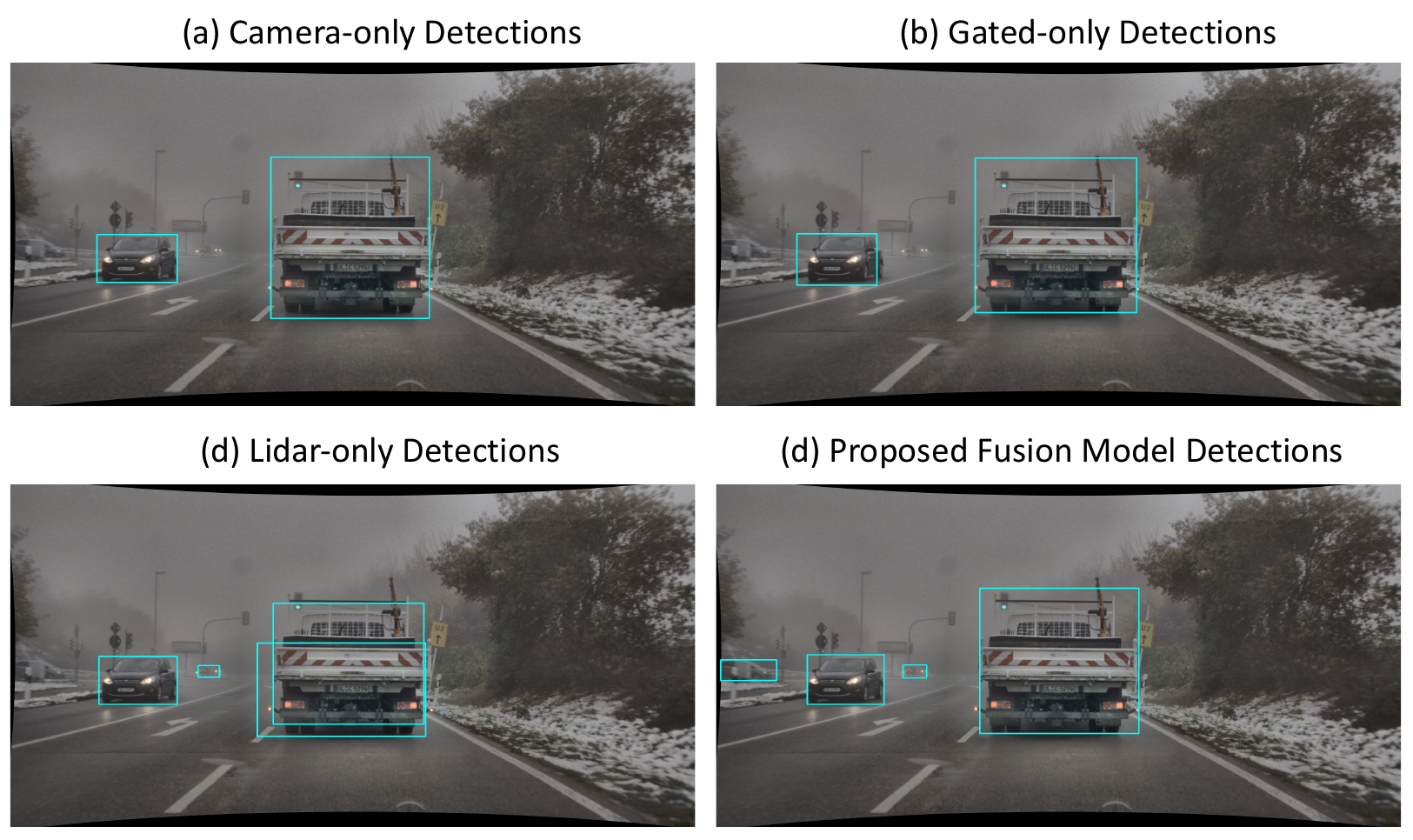}
    \vspace{-0.5em}
    \caption{The single modality model detections (Fig. 1 (a), 1 (b), and 1 (c)) failed to detect the Passenger Cars in the Adverse weather condition, which can be due to severe fog backscatter. The proposed Global Local Attention framework (Fig. 1 (d)) adapts to adverse weather by collaborative learning from Global and Local Attention features using the camera, gated, and lidar modalities.}
    \vspace{-1em}
\end{figure}


Most existing multimodal fusion methods \cite{pfeuffer2018optimal, tudela2021design, sharma2019lateral, chen2017multi} use simple concatenation or element-wise addition methods to fuse multiple modalities for object detection. These methods fail to adjust the focus of different sensors to deal with the dynamic conditions in adverse weather. In addition, the existing works \cite{zhang2020hybrid, li2020object, pfeuffer2018optimal, tudela2021design, sharma2019lateral, chen2017multi} use early or late-stage multimodal fusion and only consider low-level or high-level features from different sensors, which can be insufficient to deal with complex adverse weather conditions. For example, the scales of the objects vary during object detection. Using only local information (low-level features) or global information (high-level features) does not allow the model to fuse multimodality data in different object scales, which, however, is critical in adverse weather conditions.
 
To address the above-mentioned challenges, we leverage the attention method to adjust the focus of different sensors to deal with the dynamic adverse weather conditions. Additionally, we sequentially extract the local and global information from multiple sensors to adjust to the complex adverse weather conditions. We propose a Global-Local Attention (GLA) framework to adaptively fuse data from multiple sensors in two stages. In particular, the GLA framework learns the local and global information using the Local Attention Network and Global Attention network and thus adaptively leverages the best camera, gated, and lidar features to deal with specific weather conditions. Each Attention Network learns to allocate higher weights to the modality with better detection features. To evaluate the performance of the GLA framework, we conduct experiments on the DENSE Dataset \cite{bijelic2020seeing}. We compare our GLA framework with the state-of-art research~\cite{bijelic2020seeing, tudela2021design}. To further analyze the key components in GLA, we investigate the impact of the number of partitions in the Global Attention Network and the impact of using only Global or Local Attention.

Our main contributions are as follows.
\begin{itemize}
\item We propose a Global-Local Attention (GLA) framework to tackle object detection in adverse weather using the camera, gated, and lidar sensors to perform the dynamic fusion.

\item We perform fusion at the two stages in our proposed GLA framework. First, the Local Attention features are used as an early-stage fusion, and second, the Global Attention features as a later stage fusion method. The collaborative learning from local and global features makes our model adaptive to deal with the highly-variable adverse weather conditions.  
\item We rigorously evaluate the proposed GLA framework. The proposed GLA framework achieves an average improvement of 19.83\% mAP and 12.645\%  mAP compared with the state-of-the-art SSD-VGG model \cite{tudela2021design} and SSD Deep Entropy fusion model \cite{bijelic2020seeing}.

\item We provide an ablation study to investigate the effectiveness of the attention map and the partition number's impact. We further demonstrate the effectiveness of Global-Local attention by comparing it with the global or local only attention methods.

\end{itemize}

\section{Related Work}

\subsection{Multimodal Fusion}
Multimodal fusion aims to provide robust prediction in object detection using multi-modality inputs.
Simon \etal \cite{sharma2019lateral} proposed concatenation or element-wise addition for fusing low-level radar and camera features. Similarly, Xiaozhi \etal \cite{chen2017multi} proposed the concatenation for early/late-stage fusion and element-wise addition for the deep fusion of lidar and camera features in the architecture. 
Recently, attention-based fusion methods have attracted great interest in multimodal fusion.
Zhang \etal ~\cite{zhang2020hybrid} proposed a hybrid attention-aware fusion network (HAFNet) based on a cross-modal attention mechanism for HRI and Lidar features fusion using ATT-AFBlock. They used a global pooling layer followed by several FC layers and a sigmoid layer to perform fusion by interpreting channel-wise sigmoid weights of each modality. Dai \etal ~\cite{dai2021attentional} introduced the idea of attentional feature fusion for fusing single modality features of different scales using sigmoid layers in the Global and Local Attention aggregation for a classification task. In comparison to their method, the focus of our paper is to perform object detection in adverse weather by performing multimodal adaptive fusion by considering both local and global features. In the proposed Global Attention Network, we extract global partition-level features, i.e., global features for different partitions of an image, instead of focusing on image-level global features. Li \etal~\cite{li2020object} integrated Global Attention with the YOLOv3 model for object detection. The attention block included the Global Max Pooling and Global Average Pooling layers to generate two attention tensor types and concatenate them before passing through FC layers and the softmax layer to map the output to a probability distribution. However, the global max pooling or global average pooling extracts only image-level features instead of local and partition-level
features, which can be insufficient for detecting objects in complex adverse weather conditions.

\subsection{Object Detection in Adverse Weather}
The major challenge in the multimodal fusion for object detection in adverse weather is developing a fusion method that can adapt to highly-variable adverse weather conditions. Mees \etal \cite{mees2016choosing} showed that a novel mixture of deep network experts could be used to adaptively weight the prediction of different modality classifiers in harsh light conditions. Pfeuffer and Dietmayer \cite{pfeuffer2018optimal} focused on finding the optimal fusion method between early, mid, and late fusion methods for object detection in adverse weather and used concatenation to fuse the camera and lidar features. Moreover, Debrigode and Guillem \cite{tudela2021design} developed the SSD-VGG halfway fusion model on a Dense dataset for Adversarial weather conditions. They used a simple concatenation method to fuse the Camera and Gated features. However, these studies use the conventional fusion method, such as concatenation that cannot adapt to dynamic adverse weather conditions. 
Additionally, early or late fusion allows the model to consider only low-level features or high-level features during fusion. Bijelic \etal ~\cite{bijelic2020seeing} focused on weighted continuous multimodal fusion by calculating a multiplication matrix using entropy on inputs to scale each sensor based on available information. Their method performed an adaptive fusion of camera, gated, and lidar sensor features but lacked consideration of local (low-level) and global (partition-level) attention features during fusion. Therefore, our proposed GLA framework aims to learn from local and global features that can help detect partially visible objects in adverse weather conditions.

\begin{figure*}[!t]
    \centering
    \includegraphics[width=0.8\linewidth]{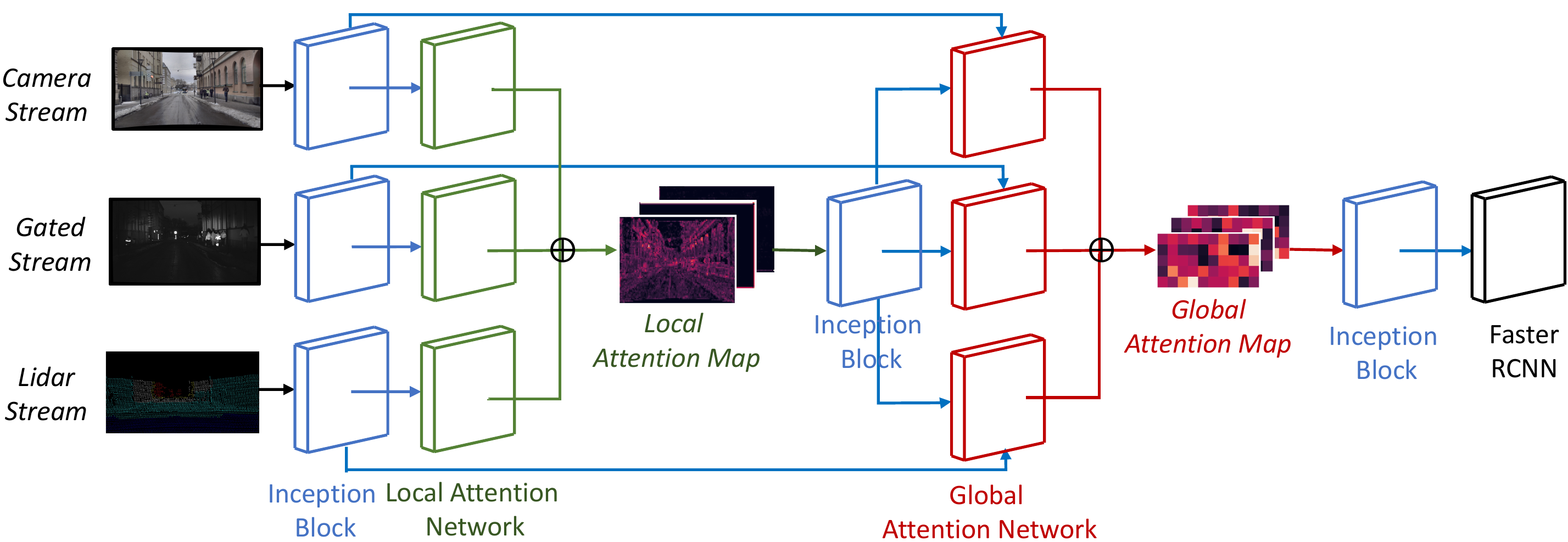}
    \caption{The proposed Global Local Attention framework consists of three branches: camera, gated, and lidar streams. For each stream, the Local Attention Network \textit{(green)} and Global Attention Network \textit{(red)} take input from inception blocks \textit{(blue)} which are either Camera, Gated, or Lidar features. The Global Attention Network has an additional input as a local attention feature. The Local/Global Attention Map is derived from the Local/Global Attention Network to perform multimodal weighted fusion represented with $\bigoplus$. Finally, the FasterRCNN analyzes the fused feature maps for object detection and outputs the final prediction.}
\end{figure*}

\section{Global-Local Attention Framework}
\subsection{Overview}

The proposed Global Local Attention framework depicted in Figure 2 consists of three branches corresponding to the camera, gated, and lidar streams. The input is processed using the Inception block for each stream, then the Local Attention Network calculates local attention feature weights and performs the first-stage multimodal fusion. Further, the Global Attention Network takes Inception block processed input (same as Local Attention Network) and the local attention feature map from first-stage fusion after processing with Inception blocks to perform the second-stage multimodal fusion. The final feature map is sent to the FasterRCNN's Region Proposal Network \cite{ren2015faster}, a widely used object detection framework. The Region Proposal Network generates anchors of different sizes and scales. Finally, the object classification and bounding boxes are generated after performing non-max suppression on these anchors.

\subsection{Local Attention Network} 
We develop a Local Attention Network to select the best local features from different modality inputs representing each pixel. 
Given an intermediate Camera, Gated, Lidar features (extracted using the first Inception block) $\mathcal{\mathbf{F_{CA}}}$, $\mathcal{\mathbf{F_G}}$, $\mathcal{\mathbf{F_L}}$ $\in \mathbf{R^{H \times W \times C}}$
with $H \times W$ feature map and $C$ channels. 
The local attention weight (LA) for camera, gated, and lidar streams can be computed as:
\begin{equation}
\label{eq:local_attention1}
\mathbf{LA_{\textbf{I}}} = \mathbf{Conv}(\mathbf{BN}(\mathbf{Conv}(\mathbf{\sigma}(\mathbf{BN}(\mathbf{Conv}(\mathcal{\mathbf{F_{I}}})))))),
\end{equation}
where $\mathbf{F_I} = \mathbf{IB(X)}$, IB denotes Inception Block, X denotes input feature for camera, gated, and lidar streams, $\mathcal{\mathbf{LA_I}}$ $\in \mathbf{R^{H x W x C}}$ denotes the local attention weights for camera ($LA_{CA}$), gated ($LA_{G}$), and lidar ($LA_{L}$) streams, $\mathcal{\mathbf{F_{I}}}$ represents corresponding camera ($F_{CA}$), gated ($F_{G}$), and lidar ($F_{L}$) intermediate features. Conv denotes conv2d layer, BN denotes Batch Normalization layer, and $\sigma$ denotes ReLU activation function. 
We then apply a channel-wise softmax operation to derive the local attention weights for the camera, gated, and lidar streams.
Leveraging the local attention weights, the first-stage multimodal feature fusion calculates a local attention feature map $F1$ to select the best feature for each pixel from different modality inputs. The calculation is summarized as follows: 
\begin{equation}
\mathbf{LA^{'}_{CA}}, \mathbf{\mathbf{LA^{'}_{G}}}, \mathbf{\mathbf{LA^{'}_{L}}} = \mathbf{Softmax(LA_{CA}, LA_{L}, LA_{G})}
\end{equation}
\begin{equation}
    \mathbf{F1} = \mathcal{\mathbf{F_{CA}}}*\mathbf{LA^{'}_{CA}} + \mathcal{\mathbf{F_{G}}}*\mathbf{LA^{'}_{G}} + \mathcal{\mathbf{F_{L}}}*\mathbf{LA^{'}_{L}}.
\end{equation}

\subsection{Global Attention Network} 
We develop a Global Attention Network, which performs partitions of an input feature to select the best global features from different modality inputs for each partition.

Moreover, given the camera stream, gated stream, lidar stream features (extracted using the first Inception block) $\mathcal{\mathbf{F_{CA}}}$, $\mathcal{\mathbf{F_G}}$, $\mathcal{\mathbf{F_L}} \in \mathbf{R^{H \times W \times C}}$ with $H \times W$ feature map and C channels. In the proposed Global Network, we partition an input feature among 5 rows and 10 columns, resulting in a total of 50 partitions before using the average pooling layer over each partition to focus on the features of partition-level features instead of image-level features. Similar to Local Attention Network, we use a conv2d, batch normalization layers, followed by ReLU activation to extract features from inputs. 
\begin{equation}
\mathbf{GA_{\textbf{I}}} = \mathbf{Conv}(\mathbf{BN}(\mathbf{Conv}(\mathbf{\sigma}(\mathbf{BN}(\mathbf{Conv}(\mathbf{AP}(\mathcal{\mathbf{F_{I}}}))))))),
\end{equation}
where $\mathcal{\mathbf{GA_I}}$ $\in \mathbf{R^{H x W x C}}$ denote the global attention feature weights for camera ($GA_{CA}$), gated ($GA_{G}$), and lidar ($GA_{L}$) streams, $\mathcal{\mathbf{F_{I}}}$ represents corresponding camera ($F_{CA}$), gated ($F_{G}$), and lidar ($F_{L}$) intermediate Features. 
Further, we apply channel-wise Softmax to compute the global attention weight (GA) for camera, gated, and lidar streams as follows:
\begin{equation}
\mathbf{\mathbf{GA^{'}_{CA}}}, \mathbf{\mathbf{GA^{'}_{G}}}, \mathbf{\mathbf{GA^{'}_{L}}} = \mathbf{Softmax(GA_{CA}, GA_{G}, GA_{L})}
\end{equation}

\begin{table*}
\caption{The performance comparison of baseline models with the proposed Global-Local Attention method over PassengerCar class. We report mAP (\%) on the DENSE dataset over easy (E), moderate (M), and hard (H) difficulty levels.}
\begin{tabularx}{\textwidth}{@{}l*{14}{C}c@{}}
\toprule
& & \multicolumn{12}{c}{\textbf{Weather Conditions}}\\
\textbf{Models} & \textbf{Day/ Night} & \multicolumn{3}{c}{\textbf{Clear}}  & \multicolumn{3}{c}{\textbf{LightFog}} & \multicolumn{3}{c}{\textbf{DenseFog}}  & \multicolumn{3}{c}{\textbf{Snow}}   \\
& & \textbf{E (\%)} & \textbf{M (\%)} & \textbf{H (\%)} & \textbf{E (\%)} & \textbf{M (\%)} & \textbf{H (\%)} & \textbf{E (\%)} & \textbf{M (\%)} & \textbf{H (\%)} & \textbf{E (\%)} & \textbf{M (\%)} & \textbf{H (\%)} \\
\midrule
\multirow{ 2}{*}{Camera} & Day   & 92.23	& 57.17	& 38.76	& 97.86	& 62.37	& 20.62	& 90.37	& 22.99	& 35.11	& 92.78	& 51.40	& 36.82 \\
& Night	& 87.49	& 49.02	& 36.45	& 91.57	& 56.11	& 45.92	& 90.28	& 52.56	& 22.03	& 94.54	& 48.55	& 29.96 \\
\multirow{ 2}{*}{Gated} & Day   & 86.81	& 55.52	& 14.78	& 95.12	& 55.52	& 14.78	& 91.55	& 31.35	& 22.67	& 91.00	& 49.84	& 32.03 \\
& Night	& 88.24	& 44.63	& 41.26	& 88.24	& 44.63	& 41.26	& 86.42	& 24.39	& 18.37	& 88.52	& 34.15	& 29.06 \\
\multirow{ 2}{*}{Lidar} & Day   & 85.62	& 46.13	& 27.55	& 97.65	& 59.16	& 16.97	& 94.42	& 23.31	& 39.25	& 94.43	& 55.37	& 31.59 \\
& Night	& 88.63	& 46.81	& 33.47	& 92.62	& 55.75	& 39.83	& 90.12	& 48.35	& 14.23	& 94.29	& 47.52	& 25.87 \\
\multirow{ 2}{*}{Camera-Gated} & Day   & \textbf{95.40}	& 71.79	& 54.22	& 98.65	& 71.75	& 39.95	& \textbf{97.90}	& 53.12	& 57.83	& \textbf{96.97}	& \textbf{70.44}	& 55.62 \\
& Night	& 93.69	& 63.26	& 52.18	& 96.62	& \textbf{69.53}	& 60.17	& \textbf{94.63}	& 63.11	& 38.23	& 96.39	& \textbf{67.74}	& 50.55 \\
\multirow{ 2}{*}{Camera-Lidar} & Day   & 95.33	& 72.26	& \textbf{55.21}	& 99.03	& 77.38	& \textbf{41.78}	& 97.86	& 48.08	& 56.55	& 96.22	& 68.97	& 57.40 \\
& Night	& \textbf{94.19}	& 63.43	& 53.37	& 96.82	& 66.36	& 65.25	& 93.38	& \textbf{63.53}	& 37.62	& 96.43	& 66.44	& 50.73 \\
\multirow{ 2}{*}{Camera-Gated-Lidar (Concat)} & Day   & 93.87	& 63.96	& 43.63	& 97.75	& 60.77	& 22.32	& 96.79	& 36.35	& 39.70	& 95.66	& 62.09	& 46.73 \\
& Night	& 91.35	& 56.42	& 47.47	& 95.12	& 60.43	& 55.81	& 93.32	& 47.87	& 24.40	& 95.47	& 57.32	& 36.54 \\
\multirow{ 2}{*}{\textbf{Proposed Method}} & Day   & \textbf{95.40}	& \textbf{73.09}	& \textbf{55.00*}	& \textbf{99.05}	& \textbf{79.69}	& \textbf{40.85*}	& 97.82	& \textbf{56.47}	& \textbf{58.16}	& \textbf{96.86*}	& 68.39	& \textbf{53.65} \\
& Night	& 93.47	& \textbf{63.44}	& \textbf{54.15}	& \textbf{97.12}	& \textbf{68.11*}	& \textbf{66.67}	& \textbf{94.36*}	& 60.61	& \textbf{43.96}	& \textbf{97.48}	& \textbf{67.28*}	& \textbf{51.19} \\
\bottomrule
\end{tabularx}
\smallskip

\textbf{*} represents the second-best model results for the specified Day/Night weather condition.
\vspace{-1em}
\end{table*}

Finally, the second-stage fusion was performed using $\mathcal{\mathbf{F1'}}$, derived from processing first-stage feature fusion $F1$ with inception block and Global Attention Weights ($GA^{'}_{CA}, GA^{'}_{G}, GA^{'}_{L}$).
The global feature map $F2$ is calculated as follows. 
\begin{equation}
    \mathbf{F2} = \mathcal{\mathbf{F1'}}*\mathbf{GA^{'}_{CA}} + \mathcal{\mathbf{F1'}}*\mathbf{GA^{'}_{G}} + \mathcal{\mathbf{F1'}}*\mathbf{GA^{'}_{L}},
\end{equation}
where $\mathbf{F1'} = \mathbf{IB(F1)}$, IB denotes Inception Block and F1 denotes first-stage feature fusion. 
The proposed fusion model for each partition assigns higher weights to the modality with better features and the weights for each corresponding partition in the global attention map sum up to one. 


\section{Evaluation}

\subsection{Dataset Description}
We evaluate the proposed GLA framework on the DENSE dataset \cite{bijelic2020seeing}. 
The dataset contains multiple modality inputs, including camera, gated, and lidar sensors for clear and adverse weather conditions such as light fog,  dense fog, and snow. 
Following \cite{bijelic2020seeing}, we project the three-dimensional lidar data to front-view considering depth, height, and pulse-intensity values.
The original Gated images in the DENSE dataset of dimension $1280\times 768$
were projected to the RGB camera plane by adding black padding of 210, 46, 280, and 360 to the top, bottom left, and right borders, respectively.

In our work, we considered PassengerCar, Pedestrian, LargeVehicle, and Ridable Vehicle classes among the total 14 classes available in the dataset, similar to \cite{tudela2021design}. We constructed the training dataset consisting of samples from all weather conditions. For clear weather samples, the information about the training and test dataset samples was taken from the GitHub repository \cite{bijelic2020seeing}. For other weather conditions, we randomly split the available samples in the training and test datasets such that around 20 percent of Day and Night samples are in the test dataset.

\subsection{Experimental Settings}
The proposed GLA method is a general framework for object detection in adverse weather, which can be used with both single-stage and two-stage object detection models. In this work, we have used the GLA framework in conjunction with the Faster RCNN model (one of the most commonly used object detection models) and implemented using Tensorflow Object Detection API. We used the FasterRCNN-InceptionV2 pre-trained model weights on the COCO dataset to perform training with a constant learning rate of 0.00002 using the Adam optimizer by setting a batch size to 2. We use the Weighted Smooth L1 Localization Loss (Huber Loss) function for both the first and second stages of the FasterRCNN model. However, the Classification Loss Function was changed from Softmax Cross-Entropy in the first stage to Sigmoid Focal loss in the second stage to address the class imbalance problem in the dataset. 
We also used Non-Maximum Suppression (NMS) \cite{hosang2017learning} to filter the predicted bounding box above the 0.7 IoU threshold. Finally, the performance evaluation of the proposed and baseline models was conducted using mean average precision (mAP) with IoU 0.5 using the Pascal VOC Detection Metrics across easy (E), moderate (M), and hard (H) difficulty cases following \cite{geiger2012we}.

\begin{figure*}[!t]
    \centering
    \includegraphics[width=17cm, height=7cm]{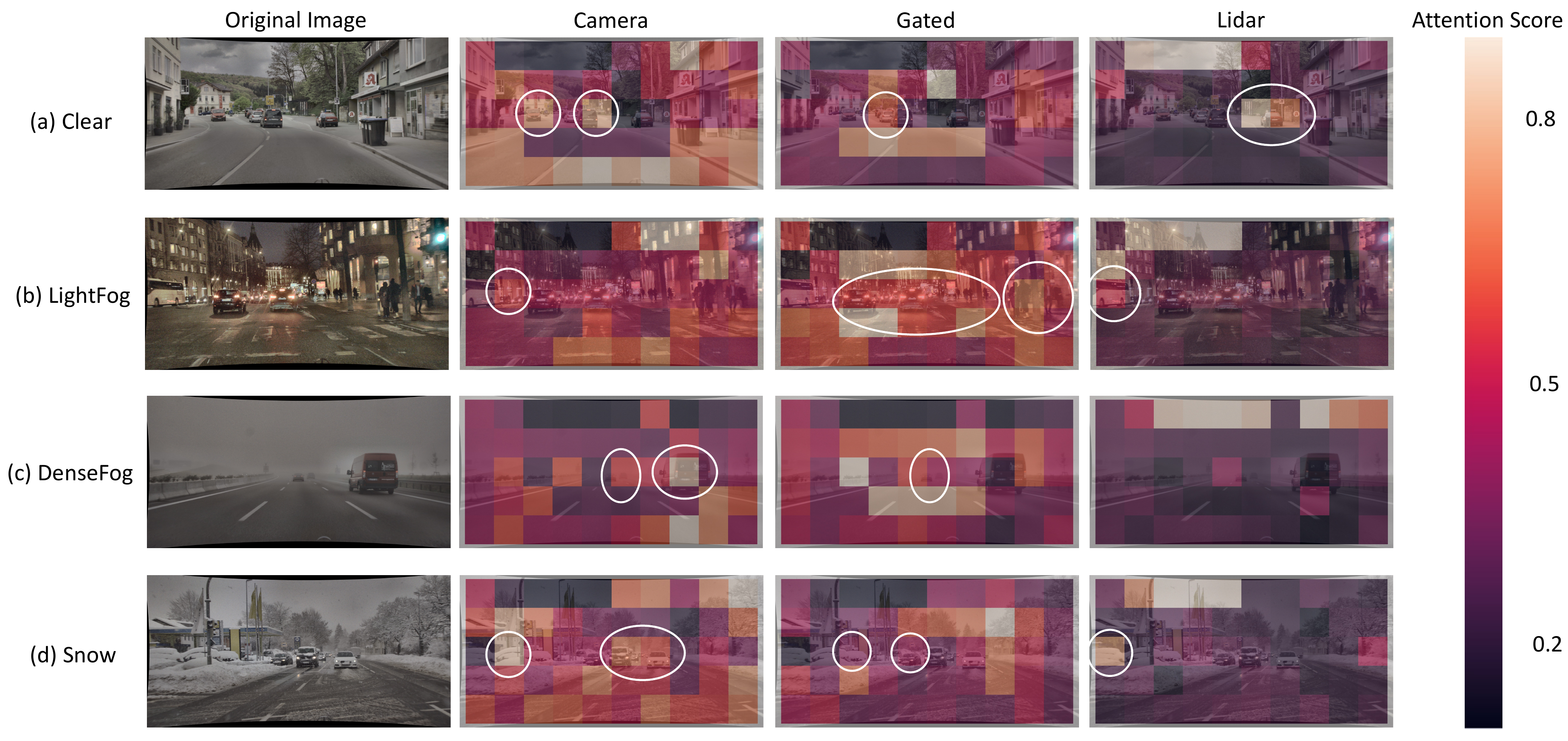}
    \caption{The camera, gated, lidar attention maps demonstrate the effectiveness of our proposed GLA framework in the clear weather (Fig. 3(a)) and adverse weather conditions, light fog (Fig. 3(b)), dense fog (Fig. 3(c)), snow (Fig. 3(d)). The attention maps show the attention score corresponding to different colors in the color map. The \textit{white} Ellipse in the Attention map highlights each modality's best detection features (i.e., higher attention score features representing PassengerCar, Pedestrian, LargeVehicle, or Ridable Vehicle).}
    \vspace{-1em}
\end{figure*}

\subsection{Evaluation Results}
This section evaluates our proposed fusion model on the individual test dataset for each weather. Similar to the state-of-art method \cite{bijelic2020seeing}, we evaluated the proposed GLA framework over PassengerCar class because these are the most prevalent class in the DENSE dataset.

We compare our proposed GLA framework with previous works \cite{bijelic2020seeing, tudela2021design} on the DENSE dataset, consisting of Day/Night samples for Clear, light fog, dense fog, and snow weather conditions. The study \cite{bijelic2020seeing} showed the superiority of their method over \cite{tzeng2017adversarial, hoffman2018cycada, qi2018frustum, ku2018joint} object detection methods for adverse weather on the DENSE dataset. Hence, we also show the effectiveness of our proposed GLA framework for adverse weather over \cite{tzeng2017adversarial, hoffman2018cycada, qi2018frustum, ku2018joint} methods by comparing it with \cite{bijelic2020seeing}.

We perform a comparison of the proposed GLA framework with three single modality baseline models: Camera-Only, Gated-Only, Lidar-Only, and two fusion baseline models: Camera-Gated, Camera-Lidar on the test dataset. 
Also, we compare the proposed GLA framework with the simple concatenation model, which has the same architecture except using concatenation to perform fusion instead of Global-Local Attention. Table I presents the results of our proposed method and other associated baseline model results over PassengerCar class. The single modality models were trained with training hyperparameters following Kitti Config in TensorFlow Object Detection API, and two baseline fusion models were trained with similar training hyperparameters as the proposed fusion model. 

The camera-only model performed best across all weather conditions among the single modality models. The proposed GLA framework has an average performance improvement of 13.72\% over the camera-only model and an average performance improvement of 8.79\% over a simple concatenation-based method keeping the modalities same; this shows the effectiveness of our proposed method.
The adaptive feature fusion of camera, gated, and lidar modalities using Local Attention and Global Attention at two stages helps perform collaborative learning from each sensor by feature exchange. The proposed GLA framework, other than single modality models, also mostly outperformed or has the second-best results compared to the camera-gated and camera-lidar baseline models. 

\begin{table}[!tb]
\caption{Performance comparison of proposed GLA framework with the state-of-art methods in terms of mAP (\%) over PassengerCar class.}
\small 
\setlength{\tabcolsep}{2pt} 
\begin{tabularx}{\columnwidth}{@{} l *{4}{C} c @{}}
    \toprule
     \textbf{Model} &\textbf{ Day/Night} & \multicolumn{4}{c@{}}{\textbf{Weather Conditions}} \\
    & \multirow{ 2}{*}{} & \textbf{Clear} & \textbf{LightFog} & \textbf{DenseFog} & \textbf{Snow} \\
    & & (\%) & (\%) & (\%) & (\%) \\
    \midrule
    \multirow{ 2}{*}{SSD-VGG \cite{tudela2021design}}   & Day    & 56.09	& 67.18	& 69.71	& 58.61 \\
    & Night    & 58.60	& 61.01	& 70.18	& 62.62 \\
    SSD Deep \cite{bijelic2020seeing}    & Day    & 67.13	& 75.47	& 74.75	& 69.24 \\
    Entropy Fusion & Night    & 62.20	& 71.96	& 68.72	& 72.04 \\
    \multirow{ 2}{*}{\textbf{Proposed Method}}    & Day    & \textbf{79.14}	& \textbf{87.92}	& \textbf{87.99}	& \textbf{78.79} \\
    & Night    & \textbf{78.32}	& \textbf{84.50}	& \textbf{82.90}	& \textbf{83.08} \\
    \bottomrule
\end{tabularx}
\vspace{-2em}
\end{table}

Table II compares the proposed GLA framework with the state-of-art multimodal fusion models SSD Deep Entropy Fusion \cite{bijelic2020seeing} and SSD-VGG \cite{tudela2021design} over PassengerCar class. In the study \cite{tudela2021design}, the proposed SSD-VGG model's overall difficulty evaluation in different weather conditions is conducted instead of individual difficulty evaluation with easy, moderate, and hard weather conditions. So, we also compare the overall difficulty evaluation results of our proposed GLA framework with the state-of-art models used in this study. Overall, our proposed GLA framework outperformed the SSD-VGG \cite{tudela2021design} model by a margin of 23.05\%-19.72\%, 20.74\%-23.49\%, 18.28\%-12.72\%, and 20.18\%-20.46\% mAP during Day-Night time for clear, light fog, dense fog, and snow weather, respectively. For reproducing the SSD Deep Entropy Fusion \cite{bijelic2020seeing} model, as the authors have not released their model codes, experimental settings, and training hyperparameters, we follow the experimental settings of this study. To perform a fair comparison, we reproduce their fusion model logic using SSD-Inception and perform training using a pretrained model since we have also used an inception feature extractor and a pretrained model for training.
\textit{Compared with their approach, the proposed GLA framework improved the object detection mAP by 12.04\%-16.12\%, 12.45\%-12.54\%, 13.24\%-14.18\%, and 9.55\%-11.04\% mAP during Day-Night time for clear, light fog, dense fog, and snow weather, respectively.
}

\section{Ablation Study}
This section provides an ablation study to investigate the effectiveness of the attention map and the impact of partition numbers in the Global Attention Network. We further demonstrate the effectiveness of Global-Local attention by comparing it with the global or local only attention methods.
\subsection{Attention Maps.}

To demonstrate the effectiveness of our proposed model in Adverse weather conditions, we visualize the camera, gated, and lidar attention map using the heatmap in Fig. 3 over clear, light fog, dense fog, and snow weathers. In the attention maps, light color represents high attention weights (0.8), violet color represents moderate attention weights (0.5), and dark color represents low attention weights (0.2). 
Our method builds attention weights for 50 partitions of an image, where different modalities can get attended for a different partition of an image. This approach makes our method adaptive to adverse weather conditions, where different sensors can be preferred depending on weather conditions.

This paragraph demonstrates the effectiveness of the proposed GLA framework in clear and light fog weather during both day and night time. 
The attention maps for a sample clear weather image in Fig. 3(a) show that the gated modality weights focused on a few distant passenger cars and the lidar attention map focused on the passenger car on the right side. At the same time, camera modality weights focused on the multiple passenger cars in two partitions of the attention map. This shows that each camera, gated, and lidar modality contributed to the passenger cars detection in the clear day weather condition. Fig. 3(b) shows the attention maps for a sample light fog weather image. Since the performance of the camera and lidar sensor can be affected during nighttime or due to fog backscatter, only a few passenger cars and large vehicles are focused on their attention map. On the other hand, the gated attention map focused on almost every center passenger car and the pedestrians walking on the right side of the road.

We also demonstrate the effectiveness of our proposed GLA framework in dense fog and snow weather conditions.
In the dense fog weather, each camera and gated attention map focused on the center distant passenger cars and a right side passenger car in Fig. 3(c). 
In the snow weather (Fig. 3(d)), the camera attention map has a higher attention weight for center passenger cars and left-side passenger cars, almost hidden in the snow. The gated and lidar sensor also contributed to detecting a few centers and left-side passenger cars. 

\subsection{Impact of the Number of Partitions in Global Attention Network.}

To study the impact of the number of partitions in our method, we construct different variants of the Global-Local Attention Networks. The Local Attention Network and all training hyperparameters were kept the same for all model variants in this experiment. The only difference is the number of partitions in the Global Network, which were set to 18 (3 rows, 6 columns, 32 (4 rows, 8 columns), 50 (5 rows, 10 columns), and 72 (6 rows, 12 columns) in the different variants.

\begin{table}[!tb]
\caption{Impact of Number of Partitions in Global Attention Network for camera-gated-lidar fusion in terms of mAP (\%) compared to proposed method over PassengerCar class.}
\small 
\setlength{\tabcolsep}{2pt} 
\begin{tabularx}{\columnwidth}{@{} l *{4}{C} c @{}}
    \toprule
     \textbf{Model} &\textbf{ Day/Night} & \multicolumn{4}{c@{}}{\textbf{Weather Conditions}} \\
    & \multirow{ 2}{*}{} & \textbf{Clear} & \textbf{LightFog} & \textbf{DenseFog} & \textbf{Snow} \\
    & & (\%) & (\%) & (\%) & (\%) \\
    \midrule
    \multirow{ 2}{*}{18 partitions}   & Day    & 78.92	& 87.10	& 86.08	& 78.50 \\
     & Night    & 78.01	& 84.58	& 81.57	& 81.42 \\
    \multirow{ 2}{*}{32 partitions}    & Day    & 79.15	& 86.68	& 86.33	& \textbf{79.40} \\
     & Night    & 78.27	& 83.97	& 82.09	& 82.62 \\
    \multirow{ 2}{*}{\textbf{50 partitions}}    & Day    & \textbf{79.15}	& \textbf{87.92}	& \textbf{87.99*}	& 78.79 \\
    \textbf & Night    & \textbf{78.32*}	& \textbf{84.50}	& \textbf{82.90}	& \textbf{83.08} \\
    \multirow{ 2}{*}{72 partitions}    & Day    & 78.93	& 85.56	& \textbf{88.42}	& 79.21 \\
     & Night    & \textbf{78.75}	& 84.39	& 80.96	& 82.90 \\
    \bottomrule
\end{tabularx}

\smallskip
\textbf{*} represents the second-best model results for the specified Day/Night weather condition.
\label{tab:partition}
\end{table}

Table~\ref{tab:partition} represents the results of four partition variants of the Global-Local Attention Network on the test dataset for overall weather conditions. In most cases, the Global Network with 50 partitions outperformed the 18 and 32 partition variants of Global Network, with exceptions for daytime snow weather cases. In this experiment, We noted the general trend of increase in mAP with an increase in partitions from 18 to 32 to 50. However, when the number of partitions in the Global Network was set to 72, the mAP mainly dropped, except for a few cases. Hence, the results suggest that the number of partitions in the Global Attention Network has an important role in working the proposed Global Local Attention model, which gives the best results with 50 partitions.

\subsection{Impact of using only Global or Local Attention method.}

Further, we investigate the impact of using Global Attention only or Local Attention only instead of the proposed Global-Local Attention method for the Adverse weather conditions. We keep the model's architecture the same except using Local Attention only for early-stage multimodal fusion or Global Attention only for late-stage multimodal fusion. 
\begin{table}[!tb]
\caption{Impact of using Global Attention only or Local Attention only for camera-gated-lidar fusion in terms of mAP (\%) compared to the proposed Global Local Attention method over PassengerCar class.}
\small 
\setlength{\tabcolsep}{2pt} 
\begin{tabularx}{\columnwidth}{@{} l *{4}{C} c @{}}
    \toprule
     \textbf{Model} &\textbf{ Day/Night} & \multicolumn{4}{c@{}}{\textbf{Weather Conditions}} \\
    & \multirow{ 2}{*}{} & \textbf{Clear} & \textbf{LightFog} & \textbf{DenseFog} & \textbf{Snow} \\
    & & (\%) & (\%) & (\%) & (\%) \\
    \midrule
    \multirow{ 2}{*}{Global Attention}   & Day    & 78.01	& 87.46	& 85.89	& \textbf{79.11} \\
     & Night    & \textbf{78.80}	& 82.47	& 81.24	& 82.56 \\
    \multirow{ 2}{*}{Local Attention}    & Day    & 78.79	& 86.23	& 86.93	& 78.64 \\
     & Night    & 78.04	& 84.02	& 82.73	& 81.69 \\
    \multirow{ 2}{*}{\textbf{Proposed Method}}    & Day    & \textbf{79.14}	& \textbf{87.92}	& \textbf{87.99}	& \textbf{78.79*} \\
    & Night    & \textbf{78.32*}	& \textbf{84.50}	& \textbf{82.90}	& \textbf{83.08} \\
    \bottomrule
\end{tabularx}

\smallskip
\textbf{*} represents the second-best model results for the specified Day/Night weather condition.
\label{table:new}
\vspace{-1em}
\end{table}

Table~\ref{table:new} summarizes the results for the Global Attention model, Local Attention model, and the proposed Global-Local Attention method. The Local Attention model showed better performance than the Global Attention model in terms of mAP for dense fog weather conditions, and the Global Attention model yielded better results for snow weather conditions. However, the superiority between the Global Attention and Local Attention model for clear and light fog weather is not clear. Notably, the proposed Global-Local Attention model outperformed the Local Attention model and Global Attention model in almost every weather condition, except daytime snow and clear nighttime weather, where the proposed method yielded second-best results. This helps to understand that although the roles of Local Attention and Global Attention are different, they are complementary and work better in combination. Hence, we propose using the Global Local Attention model for object detection in adverse weather. 


\section{Conclusion}
Recent advances in deep learning have greatly improved object detection for autonomous driving in normal weather; however, progress remains limited in adverse weather conditions. The current state-of-the-art methods are ineffective in dealing with the dynamic nature of the adverse weather by using conventional fusion methods and only considering local or global information for multimodal fusion. This work proposes Global-Local Attention (GLA), a general framework for object detection in adverse weather conditions to address these issues. The GLA framework extracts the global and local attention feature map at two stages to adaptively leverage the best of the camera, gated, and lidar features in the multimodal fusion, resulting in superior performance compared with the state-of-art methods. The proposed GLA framework outperformed the state-of-the-art research models, leading to an average improvement of 19.83\%  mAP and 12.645\%  mAP over the SSD-VGG and SSD Deep Entropy Fusion models. 

\section*{Acknowledgement}

This work is supported in part by MTU Research Excellence Fund (REF) Award.

\bibliographystyle{unsrt}
\bibliography{main}

\begin{thebibliography}{10}

\bibitem{manjunath2018radar}
Ankith Manjunath, Ying Liu, Bernardo Henriques, and Armin Engstle.
\newblock Radar based object detection and tracking for autonomous driving.
\newblock In {\em 2018 IEEE MTT-S International Conference on Microwaves for
  Intelligent Mobility (ICMIM)}, pages 1--4. IEEE, 2018.

\bibitem{minh2014feasible}
Vu~Trieu Minh and John Pumwa.
\newblock Feasible path planning for autonomous vehicles.
\newblock {\em Mathematical Problems in Engineering}, 2014, 2014.

\bibitem{sharma2019lateral}
Shobit Sharma, Girma Tewolde, and Jaerock Kwon.
\newblock Lateral and longitudinal motion control of autonomous vehicles using
  deep learning.
\newblock In {\em 2019 IEEE International Conference on Electro Information
  Technology (EIT)}, pages 1--5. IEEE, 2019.

\bibitem{zang2019impact}
Shizhe Zang, Ming Ding, David Smith, Paul Tyler, Thierry Rakotoarivelo, and
  Mohamed~Ali Kaafar.
\newblock The impact of adverse weather conditions on autonomous vehicles: how
  rain, snow, fog, and hail affect the performance of a self-driving car.
\newblock {\em IEEE vehicular technology magazine}, 14(2):103--111, 2019.

\bibitem{heinzler2020cnn}
Robin Heinzler, Florian Piewak, Philipp Schindler, and Wilhelm Stork.
\newblock Cnn-based lidar point cloud de-noising in adverse weather.
\newblock {\em IEEE Robotics and Automation Letters}, 5(2):2514--2521, 2020.

\bibitem{pfeuffer2018optimal}
Andreas Pfeuffer and Klaus Dietmayer.
\newblock Optimal sensor data fusion architecture for object detection in
  adverse weather conditions.
\newblock In {\em 2018 21st International Conference on Information Fusion
  (FUSION)}, pages 1--8. IEEE, 2018.

\bibitem{tudela2021design}
Guillem Tudela~Debrigode.
\newblock Design of an image fusion object detection algorithm for the
  autonomous driving during adversarial weather conditions.
\newblock 2021.

\bibitem{chen2017multi}
Xiaozhi Chen, Huimin Ma, Ji~Wan, Bo~Li, and Tian Xia.
\newblock Multi-view 3d object detection network for autonomous driving.
\newblock In {\em Proceedings of the IEEE conference on Computer Vision and
  Pattern Recognition}, pages 1907--1915, 2017.

\bibitem{zhang2020hybrid}
Peng Zhang, Peijun Du, Cong Lin, Xin Wang, Erzhu Li, Zhaohui Xue, and Xuyu Bai.
\newblock A hybrid attention-aware fusion network (hafnet) for building
  extraction from high-resolution imagery and lidar data.
\newblock {\em Remote Sensing}, 12(22):3764, 2020.

\bibitem{li2020object}
Wei Li, Kai Liu, Lizhe Zhang, and Fei Cheng.
\newblock Object detection based on an adaptive attention mechanism.
\newblock {\em Scientific Reports}, 10(1):1--13, 2020.

\bibitem{bijelic2020seeing}
Mario Bijelic, Tobias Gruber, Fahim Mannan, Florian Kraus, Werner Ritter, Klaus
  Dietmayer, and Felix Heide.
\newblock Seeing through fog without seeing fog: Deep multimodal sensor fusion
  in unseen adverse weather.
\newblock In {\em Proceedings of the IEEE/CVF Conference on Computer Vision and
  Pattern Recognition}, pages 11682--11692, 2020.

\bibitem{dai2021attentional}
Yimian Dai, Fabian Gieseke, Stefan Oehmcke, Yiquan Wu, and Kobus Barnard.
\newblock Attentional feature fusion.
\newblock In {\em Proceedings of the IEEE/CVF Winter Conference on Applications
  of Computer Vision}, pages 3560--3569, 2021.

\bibitem{mees2016choosing}
Oier Mees, Andreas Eitel, and Wolfram Burgard.
\newblock Choosing smartly: Adaptive multimodal fusion for object detection in
  changing environments.
\newblock In {\em 2016 IEEE/RSJ International Conference on Intelligent Robots
  and Systems (IROS)}, pages 151--156. IEEE, 2016.

\bibitem{ren2015faster}
Shaoqing Ren, Kaiming He, Ross Girshick, and Jian Sun.
\newblock Faster r-cnn: Towards real-time object detection with region proposal
  networks.
\newblock {\em Advances in neural information processing systems}, 28:91--99,
  2015.

\bibitem{hosang2017learning}
Jan Hosang, Rodrigo Benenson, and Bernt Schiele.
\newblock Learning non-maximum suppression.
\newblock In {\em Proceedings of the IEEE conference on computer vision and
  pattern recognition}, pages 4507--4515, 2017.

\bibitem{geiger2012we}
Andreas Geiger, Philip Lenz, and Raquel Urtasun.
\newblock Are we ready for autonomous driving? the kitti vision benchmark
  suite.
\newblock In {\em 2012 IEEE conference on computer vision and pattern
  recognition}, pages 3354--3361. IEEE, 2012.

\bibitem{tzeng2017adversarial}
Eric Tzeng, Judy Hoffman, Kate Saenko, and Trevor Darrell.
\newblock Adversarial discriminative domain adaptation.
\newblock In {\em Proceedings of the IEEE conference on computer vision and
  pattern recognition}, pages 7167--7176, 2017.

\bibitem{hoffman2018cycada}
Judy Hoffman, Eric Tzeng, Taesung Park, Jun-Yan Zhu, Phillip Isola, Kate
  Saenko, Alexei Efros, and Trevor Darrell.
\newblock Cycada: Cycle-consistent adversarial domain adaptation.
\newblock In {\em International conference on machine learning}, pages
  1989--1998. PMLR, 2018.

\bibitem{qi2018frustum}
Charles~R Qi, Wei Liu, Chenxia Wu, Hao Su, and Leonidas~J Guibas.
\newblock Frustum pointnets for 3d object detection from rgb-d data.
\newblock In {\em Proceedings of the IEEE conference on computer vision and
  pattern recognition}, pages 918--927, 2018.

\bibitem{ku2018joint}
Jason Ku, Melissa Mozifian, Jungwook Lee, Ali Harakeh, and Steven~L Waslander.
\newblock Joint 3d proposal generation and object detection from view
  aggregation.
\newblock In {\em 2018 IEEE/RSJ International Conference on Intelligent Robots
  and Systems (IROS)}, pages 1--8. IEEE, 2018.

\end{thebibliography}

\end{document}